\documentclass[submission,copyright,creativecommons]{eptcs}
\providecommand{\event}{ICLP 2026} 

\usepackage{amsmath}
\usepackage{amsfonts} 
\usepackage{graphicx}
\usepackage{multirow}
\usepackage{booktabs}
\usepackage{tabularx}
\usepackage{pifont}
\usepackage{algorithm}
\usepackage{algpseudocode}
\usepackage{array}
\usepackage[table]{xcolor}
\usepackage{url}
\usepackage{caption} 
\usepackage{pdfpages}
\usepackage{hyperref}

\usepackage{listings}

\usepackage{iftex}

\ifpdf
  \usepackage{underscore}         
  \usepackage[T1]{fontenc}        
\else
  \usepackage{breakurl}           
\fi

\title{Identifying Good Rules for Efficient SAT Encodings of Single-Constant Multiplication Using Machine Learning}
\author{Chufeng Jiang
\institute{The City University of New York\\ New York, USA}
\email{cjiang@gradcenter.cuny.edu}
\and
Neng-Fa Zhou
\institute{The City University of New York\\ New York, USA}
\email{nzhou@brooklyn.cuny.edu}
}
\def\titlerunning{Identifying Good Rules for SAT Encoding of SCM Using Machine Learning}
\def\authorrunning{C. Jiang and N. Zhou}
\begin{document}
\maketitle

\begin{abstract}
The Single Constant Multiplication problem is a fundamental NP-hard optimization task in hardware design, which seeks to decompose a fixed constant using only additions, subtractions, and bit-shifts. Although dynamic programming methods can produce near-optimal SAT encodings for SCM, their encoding cost remains high for large constants. We propose a neuro-symbolic framework that accelerates SCM SAT encoding by identifying \emph{good rules} for guiding operator selection during decomposition. Our approach employs a graph neural network model to predict promising operator types from constant decompositions, and exploits the resulting confidence scores to prune no-good choices in the symbolic search. Experimental results on unseen 17--32 bit constants demonstrate one to two orders of magnitude reductions in encoding time, over 97\% reduction in memory usage, and an order-of-magnitude decrease in branching, while preserving near-optimal encoding quality in terms of additions. These results show that learning-guided symbolic strategies can significantly improve the scalability and efficiency of SCM encoding. Our code and data are publicly available at:~\url{https://github.com/Chufeng-Jiang/SCM_MLDP}
\end{abstract}

\section{Introduction}

Constant multiplication by fixed coefficients, known as \emph{Single Constant Multiplication} (SCM), is a fundamental operation in digital systems, especially in digital signal processing, image processing, communication circuits, and hardware accelerators for linear transforms~\cite{potkonjak1996multiple, voronenko2007multiplierless}. When the coefficient is known at compile time, implementing the multiplication using a general-purpose multiplier is often unnecessarily expensive in terms of area, delay, and energy~\cite{gustafsson2008towards}. A standard optimization is therefore to realize the multiplication using only additions, subtractions, and bit-shifts, where shifts are typically much cheaper than arithmetic operators because they can often be implemented by wiring~\cite{goldberg1990computer}. The quality of such an implementation depends critically on how the constant is decomposed: different decompositions of the same constant can lead to substantially different numbers of adders, and thus to different circuit sizes and optimization costs.

SCM is important not only for arithmetic circuit design, but also for SAT-based optimization. An SCM constraint, of the form $c \cdot x = y$ where $c$ is an integer constant, and $x$ and $y$ are integer-domain variables, can be encoded into SAT via additions, subtractions, and shifts~\cite{dempster1994constant,bierlee2024single}. Since subtractions can be expressed as additions, and bit-shifts are free in SAT, the encoding quality is primarily determined by minimizing the number of additions (the \textit{min-k} objective) or the number of full/half adders (the \textit{min-a} objective).

In SAT encodings, each addition or subtraction is translated into boolean variables and clauses that model bit-level sum and carry propagation~\cite{metodi2013boolean}. As a result, the chosen decomposition directly determines the quality of the resulting boolean encoding. However, finding high-quality SCM decompositions is computationally challenging. The problem is NP-hard, and the number of possible decompositions grows rapidly with the magnitude of the constant, leading to an exponential increase in search time and making exhaustive approaches impractical for large instances~\cite{bernstein1986multiplication}.

Traditional approaches, such as searching directed acyclic graphs of intermediate values~\cite{tummeltshammer2007time} or applying heuristic representations like canonical signed digit forms~\cite{hewlitt2000canonical}, typically yield sub-optimal solutions. More recent methods mitigate this cost by relying on precomputed recipes of optimal SCM encodings for selected constants and bit-widths~\cite{bierlee2024single}. However, it is infeasible for encoders to carry the full range of constants encountered in practice.

Our previous work addressed this challenge by introducing a dynamic programming (DP) approach implemented with tabling in Picat~\cite{zhoupicat2026}. The method recursively decomposes a target constant $c$ using one of three operator types: \texttt{SPLUS} ($(c_l \ll s) + c_r$), \texttt{SMINUS} ($(c_l \ll s) - c_r$), or \texttt{MINUSS} ($c_l - (c_r \ll s)$), where $\ll s$ denotes a left shift by $s$ bits~\cite{bierlee2024single}. By reusing intermediate subproblems, DP substantially improves over naive exhaustive search. However, it still needs to explore all possible candidate decompositions, leading to high time and memory costs as the constant size increases.

In this paper, we propose to accelerate SCM decomposition by integrating machine learning (ML) to guide the symbolic search process. Motivated by recent work in neuro-symbolic and explainable ML for logic-based systems~\cite{padalkar2024using,mcginness2024fold}, we learn \emph{good rules} that guide operator selection during decomposition. Concretely, we represent SCM decompositions as graphs and train a Graph Neural Network (GNN) model to predict promising operator choices from decompositions. The learned predictions are used as strategies to prune unpromising branches in the DP search.

Overall, our approach combines data-driven learning with symbolic reasoning: the GNN model provides high-level learned guidance, while Picat’s logic programming framework ensures precise and exhaustive exploration of the constrained search space. This neuro-symbolic integration leverages the strengths of both paradigms, enabling more efficient and scalable solutions to the SCM problem. 

The paper is structured as follows: Section~\ref{bk} introduces the background, Section~\ref{method} details the methodologies of the ML model, Section~\ref{exp} presents the experiments and evaluation, Section~\ref{relatedwork} reviews related work, and Section~\ref{conclusion} provides the conclusion.

\section{Background}
\label{bk}

We consider the SCM of the form
\begin{equation}
c \cdot x = y,
\end{equation}
where $c$ is a fixed integer constant and $x,y$ are integer-domain variables. As any even constant can be written as $c = c' \cdot 2^s$ for some odd $c'$ and $s \ge 1$.
Without loss of generality, we restrict our attention to \emph{odd} constants $c > 1$.

In SAT encodings, multiplication by an odd constant $c$ is recursively decomposed using the following operator types:
\[
\begin{array}{r l}
\text{SPLUS:}  & c = (c_l \ll s) + c_r, \\
\text{SMINUS:} & c = (c_l \ll s) - c_r, \\
\text{MINUSS:} & c = c_l - (c_r \ll s),
\end{array}
\]
where $c_l, c_r$ are positive odd integers, $s \ge 1$, and $\ll$ denotes a left-shift~\cite{bierlee2024single}. Each application of an operator corresponds to one arithmetic operation combined with a shift. The shift-and-add algorithm only uses SPLUS, requiring $n-1$ additions, where $n$ is the number of 1 bits in binary representation.

\paragraph{Example 1.} Consider the constant multiplication $1759 \cdot x$. The binary representation is $11011011111$, which contains nine 1's, and so the shift-and-add algorithm requires eight additions, thus the cost is 8. Figure~\ref{fig:1759example} shows a more efficient decomposition.

\begin{figure}[h]
\centering

\[
\begin{aligned}
1759\cdot x &= (27\cdot x) \ll 6 + 31\cdot x,
&& \text{$1759 = 27 \cdot 2^6 + 31$},\\
27\cdot x &= 31\cdot x - x\ll 2,
&& \text{$27 = 31 - 1 \cdot 2^2$},\\
31\cdot x &= x\ll 5 - x.
&& \text{$31 = 1 \cdot 2^5 - 1$}.
\end{aligned}
\]

\caption{Example decomposition of $1759$.}
\label{fig:1759example}
\end{figure}

\noindent
The first equation decomposes $1759$ into two smaller constants, $27$ and $31$, with a left shift of $6$, while the remaining equations recursively decompose $27$ and $31$ until reaching the base constant $1$. This decomposition uses three arithmetic operations, one for each non-base constant. Since shifts are cost-free in the \textit{min-k} objective, only additions and subtractions are counted, yielding a total cost of 3.

\paragraph{Dynamic Programming Method.} 

The DP method constructs an SCM encoding by decomposing the binary representation of $c$ into bit-stream chunks using one of the operator types when $c$ is not equal to 1, and it recursively applies the method to each chunk or its complement. When multiple decompositions are possible, the DP method explores all of them~\cite{zhoupicat2026}. The corresponding Picat implementation is presented below.

\noindent
\rule{\textwidth}{1pt}
{\footnotesize
\begin{verbatim}
import ordset.

table (+, +, mmin, -)
scm(1, XBits, Cost, Plan) =>
    Cost = 0, Plan = [].
    
scm(C, XBits, Cost, Plan) =>                     
    (split_pp(C, C1, C2, S),           % SPLUS: C = (C1 << S) + C2
     OP = $splus(C, C1, C2, S)
    ;
     split_pn(C, C1, C2, S),           % SMINUS: C = (C1 << S) - C2
     OP = $sminus(C, C1, C2, S)
    ;
     split_np(C, C1, C2, S),           % MINUSS: C = C1 - (C2 << S)
     OP = $minuss(C, C1, C2, S)
    )
    comp_op_cost(XBits, OP, ThisCost),
    scm(C1, XBits, _Cost1, Plan1),
    scm(C2, XBits, _Cost2, Plan2),
    Plan12 = union(Plan1, Plan2),
    Plan = Plan12.insert((OP, ThisCost)),
    Cost = sum([T : (_, T) in Plan]).
\end{verbatim}
}
\rule{\textwidth}{1pt}

As a plan is represented as an ordered set, the program imports the \texttt{ordset} module in the beginning. The predicate \texttt{scm(C, XBits, Cost, Plan)} returns a plan (\texttt{Plan}) with cost (\texttt{Cost}) for the multiplication \texttt{C*X}, where \texttt{XBits} is the bit-width of \texttt{X}. It minimizes the number of additions when \texttt{XBits} is 0. The first rule handles the base case when \texttt{C = 1}. The second rule encodes the DP recurrences using a disjunction of three branches. After decomposing the binary representation of \texttt{C} into \texttt{C1} and \texttt{C2}, the rule recursively obtains a plan \texttt{Plan1} for \texttt{C1} and a plan \texttt{Plan2} for \texttt{C2}. It then forms a plan for \texttt{C} by taking the union of \texttt{Plan1} and \texttt{Plan2},  adding the newly derived operation together with its associated cost. The total cost is the sum of the costs of all operations in the resulting plan. Note that the table mode \texttt{mmin} is used, which instructs Picat to table all minimum-cost solutions. If the \texttt{min} table mode were used instead, the final solution might not be as optimal.

The disjunction introduces nondeterministic search: if one branch fails, the next branch is automatically tried. The branching factor is 3, and the search cost remains high for large constants. The \texttt{comp\_op\_cost(XBits, OP, Cost)} predicate binds \texttt{Cost} to the cost of operation \texttt{OP}. For the \textit{min-k} objective, the cost is always 1. For the \textit{min-a} objective, the cost is the number of adders needed for the operation, which is dependent on \texttt{XBits}. In this paper, we focus on the \textit{min-k} objective only.

\section{Identifying Good Rules Using Machine Learning}
\label{method}

This section presents our approach to learning operator-selection 
heuristics for the SCM decomposition problem. We model each 
decomposition step as a directed graph and train a GNN to predict promising 
operator types.

\begin{figure}[h]
    \centering
    \includegraphics[width=1\linewidth]{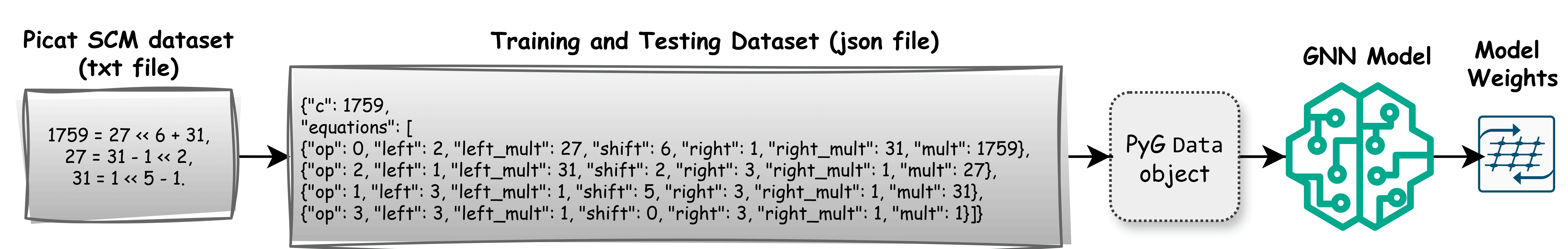}
    \caption{Training workflow of the proposed framework. DP-generated decomposition plans are converted into graph-structured data, enabling a GNN to learn operator-selection rules.}
    \label{workflowT}
\end{figure}

\begin{figure}[h]
    \centering
    \includegraphics[width=1\linewidth]{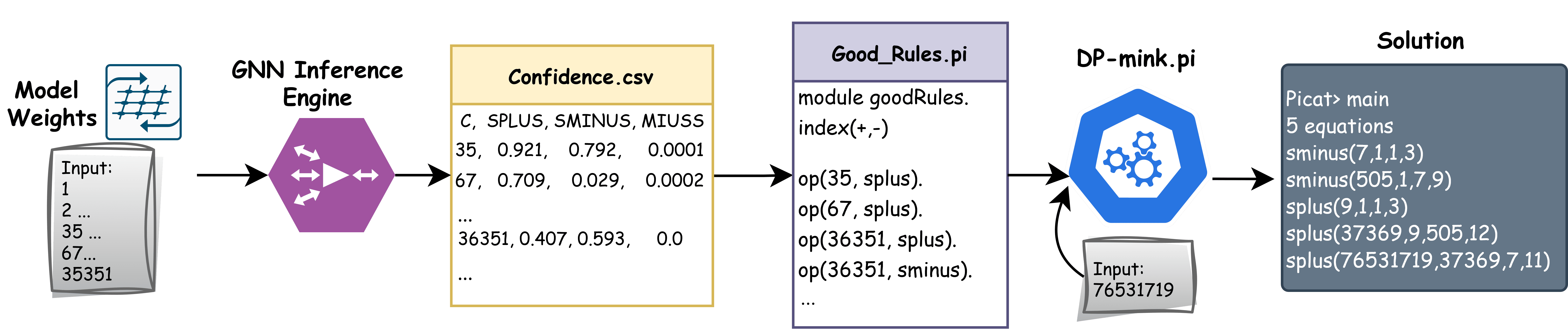}
    \caption{Inference pipeline and integration of learned rules. The trained GNN produces operator probability distributions, which are translated into high-confidence rules and incorporated into the DP solver to guide symbolic search.}
    \label{workflowI}
\end{figure}

As illustrated in Figure~\ref{workflowT}, each SCM decomposition is 
converted into a PyG data object (directed graph), where nodes correspond to decomposition 
steps and edges encode the dependency relations between subproblems 
(described in Section~\ref{graphrep}). The GNN is trained to predict 
the operator type at each decomposition step, with ground-truth labels 
derived from the decomposition plans generated by the DP method. During inference, as presented in 
Figure~\ref{workflowI}, the trained GNN outputs a probability 
distribution over the three operator types for each encountered 
constant.

\subsection{GNN Framework}

This section first describes how SCM decompositions are represented as graphs, followed by the feature design and the message-passing mechanism used for operator prediction.

\subsubsection{Graph-Based Representation for SCM Decomposition}
\label{graphrep}
To capture the evolving nature of the search process, we model each decomposition step as a directed graph $\mathcal{G} = (\mathcal{V}, \mathcal{E})$. This representation treats the SCM problem not merely as a numerical task, but as a graph-structured reasoning problem, where:

\begin{itemize}
    \item \textbf{Nodes ($\mathcal{V}$)} are a set of values, including the target and intermediate constants.
    \item \textbf{Edges ($\mathcal{E}$)} encode the operational dependencies between the node and its left and right operands.
\end{itemize}

\begin{figure}[h]
    \centering
    \includegraphics[width=1\linewidth]{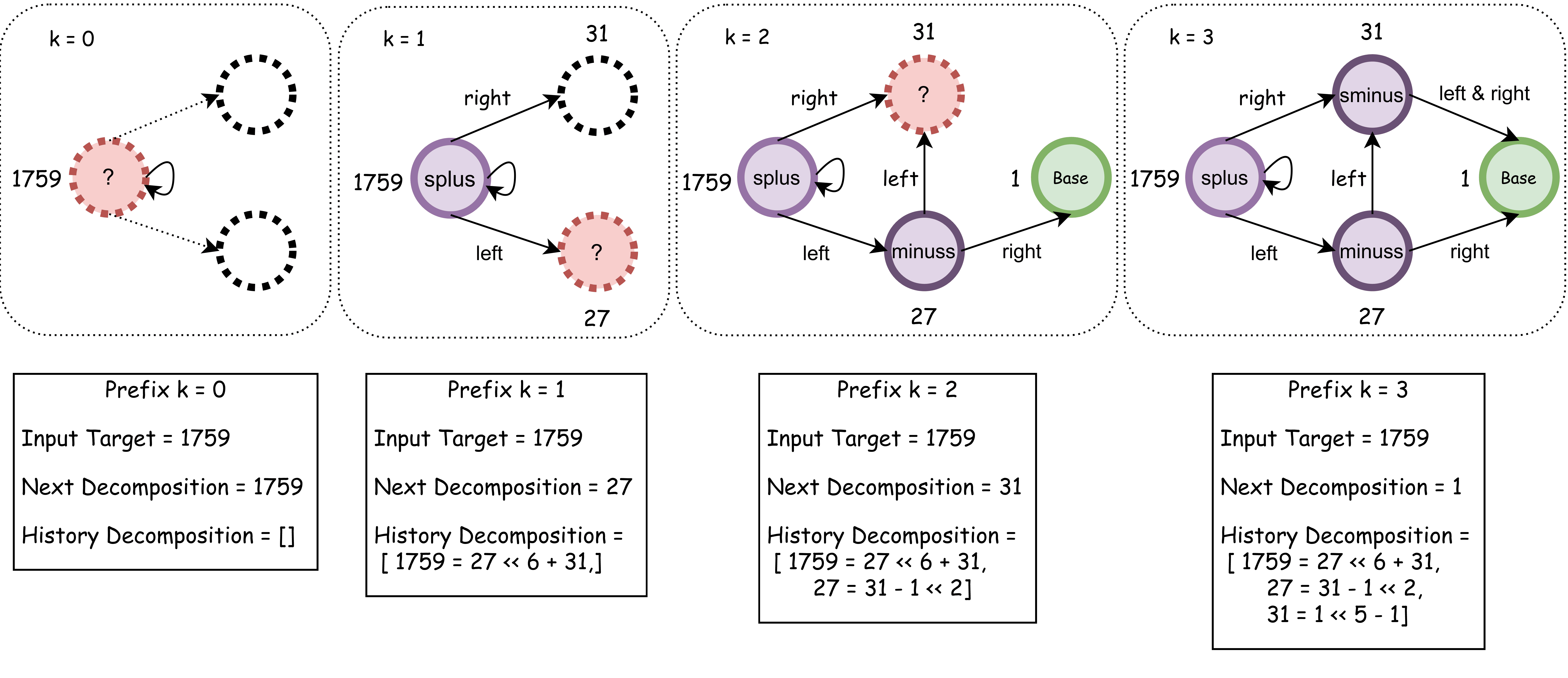}
    \caption{Graph-based representation of the decomposition process for constant 1759. Each intermediate graph, corresponding to a prefix of the decomposition, forms a training instance.}
    \label{gscm}
\end{figure}

As illustrated in Figure~\ref{gscm}, each graph instance corresponds to a decomposition step with a prefix of length $k$ (also referred to as a $k$-step history decomposition). This formulation reflects the fact that different constants admit decompositions with varying numbers of steps and intermediate subproblems. Consequently, a single SCM solution induces a sequence of graph instances, each representing the decomposition step at a particular recursion depth.

At each step, the current constant to be decomposed is designated as a \emph{prediction node} (shown as the red node), while previously resolved constants form the historical context (purple nodes). The decomposition terminates when the constant 1 is reached (green node). We explicitly model intermediate steps by associating each node requiring a decomposition decision with a corresponding partial graph, enabling the model to predict operator types incrementally rather than only at the final solution level. As a result, the GNN model is trained to perform relational reasoning by leveraging accumulated structural context while guiding the next decomposition decision.

By abstracting multipliers as nodes and decomposition relationships as edges, we combine graph-level structure with local vertex attributes. This alignment with symbolic search allows the model to function as a learned heuristic: extending the graph via a predicted operator corresponds to expanding a promising search branch (the red-node branch), while low-confidence predictions provide a principled basis for pruning unpromising paths in the DP table.

\begin{table}[ht]
\centering
\scriptsize
\caption{Selected Node and Edge Features}
\label{tab:features}
\vspace{0.5em}
\setlength{\tabcolsep}{4pt}
\renewcommand{\arraystretch}{1.18}
\begin{tabular}{p{0.28\linewidth} p{0.72\linewidth}}
\toprule
\multicolumn{1}{c}{\textbf{Features}} &  \multicolumn{1}{c}{\textbf{Description}} \\
\midrule

\multicolumn{2}{>{\columncolor{lightgray}}l}{\textit{\textbf{Node Features (subset)}}} \\
\rowcolor[gray]{0.9} \multicolumn{2}{c}{\textit{Numerical Features}} \\
\texttt{Normalized Log-space Positions} & Calculated as 
$\displaystyle \frac{\log_2(c+1)}{32}$, approximating the relative shift depth of constant. \\ \midrule

\texttt{Ones Density} &
Calculated as $ \frac{\# ones\ in\ binary\ representation}{bit\ length\ of\ c}$, evaluating the fraction of 1s in $c$. \\ \midrule

\rowcolor[gray]{0.9} \multicolumn{2}{c}{\textit{Relational Features}} \\
\texttt{Left/Right-operand Distance} &
The normalized index difference between a node and its left operand in the graph ordering, reflecting their relative positions in the decomposition sequence.
 \\ \midrule
\texttt{Left/Right-to-parent Ratio} &
This feature measures how large the left operand is relative to the parent constant. \\ \midrule
\texttt{Common 1-bit Ratio} &
 This feature counts how many bit positions are set to 1 in both the current constant $c_l$ or $c_r$ and the target $c$, indicating their similarity at the bit level. \\  \midrule

\rowcolor[gray]{0.9} \multicolumn{2}{c}{\textit{Structural Features}} \\
\texttt{Absolute DFS Level} &
The normalized index of a node in the depth-first traversal order, indicating its relative position (early vs.\ late) in the decomposition. The traversal corresponds to a DFS-style tree search over the DP method, in which $c_l$ and $c_r$ are fully explored before backtracking to their parent $c$. \\ \midrule

 \rowcolor[gray]{0.9} \multicolumn{2}{c}{\textit{Binary Indicator Features}} \\
 \texttt{Operator Type} &
One-hot over four operator classes, indicating operator types. \\ \midrule
 \texttt{Binary Representation} &
 A fixed-length binary vector encoding the bit pattern of the constant $c$. \\

\midrule
\multicolumn{2}{>{\columncolor{lightgray}}l}{\textit{\textbf{Edge Features (subset)}}} \\
\rowcolor[gray]{0.9} \multicolumn{2}{c}{\textit{Relational Features}} \\

\texttt{Is Left Operand} &
True if the edge corresponds to the left operand $c_l$ of the current constant $c$.
\\ \midrule 

\texttt{Normalized Relative Signed Distance} &
The normalized signed index difference between two connected nodes, showing whether the edge points backward or forward in the decomposition order, normalized by the graph size. \\ \midrule

\rowcolor[gray]{0.9} \multicolumn{2}{c}{\textit{Structural Features}} \\
\texttt{Normalized Absolute Index Distance} & The absolute difference between the indices of the root and child nodes in the graph ordering, normalized by the graph size.\\ \midrule

\texttt{Next-step Edge} &
Indicating currently-available nodes should be the future prediction node.\\  \midrule 

\rowcolor[gray]{0.9} \multicolumn{2}{c}{\textit{Numerical Features}} \\
\texttt{Normalized Shift Amount} &
Calculated as $\displaystyle \frac{s}{S_{\max}}$, indicating the relative magnitude of the left shift applied on the edge. \\ 

\bottomrule
\end{tabular}
\end{table}

\subsubsection{Feature Engineering}
\label{fe}

We design a comprehensive set of node and edge features to characterize each decomposition step in an instance. Each node is represented by a 199-dimensional feature vector, while each edge is encoded using a 12-dimensional vector. A subset of node and edge features is summarized in Table~\ref{tab:features}. These features can be broadly grouped into four categories: (i) numerical features capturing arithmetic properties of constants (e.g., magnitude and bit patterns), (ii) relational features describing dependencies between nodes and operands, (iii) structural features encoding the position of nodes within the decomposition process, and (iv) binary indicator features representing operator types and functional roles.

\paragraph{Example 2: Node Feature Representation.}

\begin{figure}[h]
    \centering
    \includegraphics[width=0.7\linewidth]{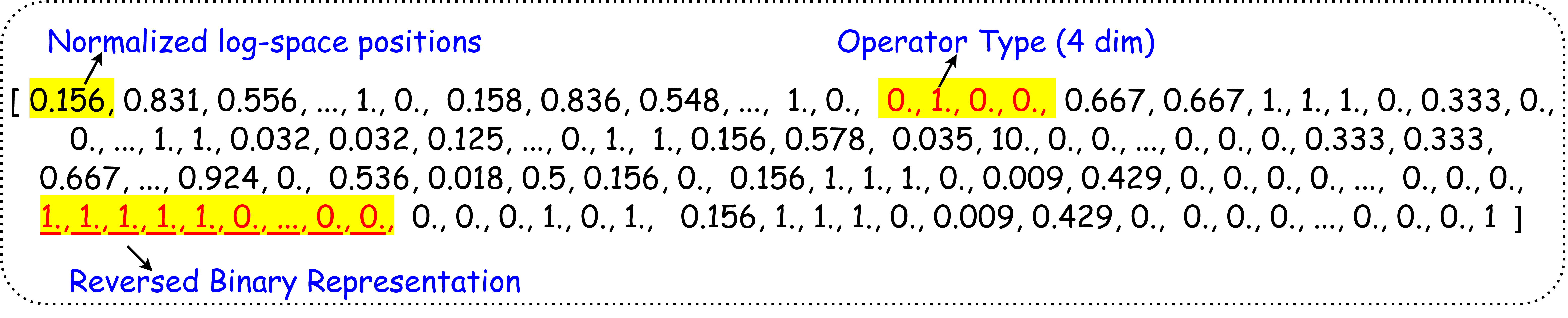}
    \caption{Node feature representation for the constant 31.}
    \label{node}
\end{figure}

As illustrated in Figure~\ref{node}, each node (e.g., Node(31)) is represented by a 199-dimensional feature vector. The first feature encodes the log-normalized magnitude of the constant, computed as $\log_2(31+1)/32 = 0.156$. The remaining features are constructed using various transformations to capture both numerical and structural properties of the node. Among these, the operator type is encoded using a one-hot vector (e.g., $[0,1,0,0]$ for the \texttt{MINUSS} operator). In addition, the 32-bit binary representation of the constant is included at the beginning of the final row of the feature segment, stored in reversed bit order. All features are concatenated to form the complete node representation.

\begin{figure}[h]
    \centering
    \includegraphics[width=0.8\linewidth]{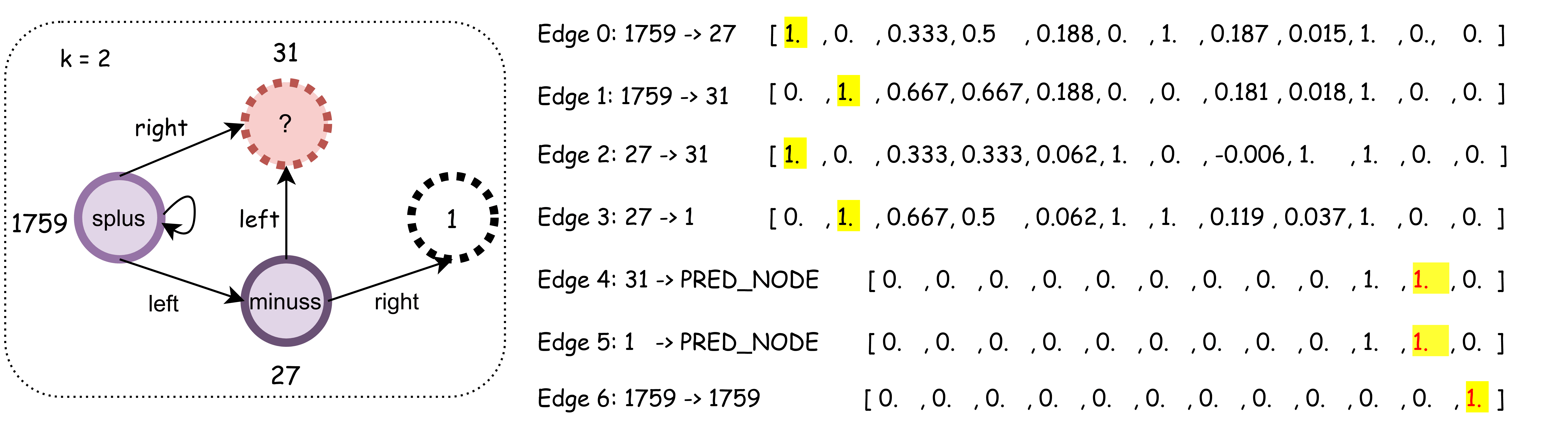}
    \caption{Edge feature representation for a graph with prefix $k=2$.}
    \label{edgek2}
\end{figure}

\paragraph{Example 3: Edge Feature Representation.}
Figure~\ref{edgek2} illustrates how relational dependencies in a graph state with prefix length $k=2$ are encoded using edge feature vectors. The first two dimensions serve as functional indicators, specifying whether the edge corresponds to a left operand ($c_l$) or a right operand ($c_r$). These are followed by numerical attributes that quantify arithmetic relationships, such as shift-related information. The second-to-last dimension encodes the structural status of the \textit{Next-step} edge, marking frontier nodes that are candidates for further expansion.

\subsubsection{Graph Neural Network Model}  
\label{mpf}

We employ a lightweight GNN that operates directly on the decomposition graphs introduced in the previous section. The GNN serves as a learned control model that scores candidate operator types based on the current decomposition step. We adopt a message-passing architecture because SCM decomposition graphs are irregular, dynamically growing, and vary substantially in size and topology across instances. Message-passing GNNs naturally accommodate these properties and satisfy key invariance requirements, namely permutation invariance and equivariance, ensuring that predictions are insensitive to the ordering or naming of intermediate constants~\cite{keriven2019universal}.

Given a constant as input, the model transforms it into a graph PyG data object for training and produces a probability distribution over the operator types. The architecture consists of four main components:
\begin{enumerate}
    \item \textit{Input Projection.} Each node feature vector $x_i \in \mathbb{R}^{199}$ is projected into a shared hidden space of dimension $d$ using a linear layer. This allows heterogeneous symbolic features to be embedded into a common latent representation.
    \item \textit{Message-Passing Layers.} A stack of $N$ graph attention layers performs iterative message passing over the decomposition graph. Edge attributes are incorporated directly into the attention mechanism, enabling the model to condition information flow on operand roles, shift values, and structural edge types.
    \item \textit{Normalization and Residual Connections.} Each message-passing layer is followed by graph normalization and a residual connection. Graph normalization stabilizes training across graphs of different sizes, while residual connections preserve low-level arithmetic information and improve gradient flow in deeper networks.
    \item \textit{Global Pooling and Prediction Head.} Node representations are aggregated via global mean pooling to obtain a graph-level embedding. This embedding is passed to a multi-layer perceptron that produces logits over operator types.
\end{enumerate}

The model predicts only the operator type for the current decomposition step. This design keeps the learning task focused while preserving the soundness and correctness guarantees of the underlying symbolic encoding.

We adopt a GNN instead of an MLP or CNN because SCM decomposition is inherently relational and history-dependent. Graph representations naturally capture dependencies among intermediate constants and subexpressions, while message passing enables past subproblems to inform future decisions, aligning with DP search. Global mean pooling suffices as predictions depend on the overall state and should be invariant to graph size and node ordering.

\subsubsection{Message Passing Semantics}

\begin{figure}[ht]
    \centering
    \includegraphics[width=0.8\linewidth]{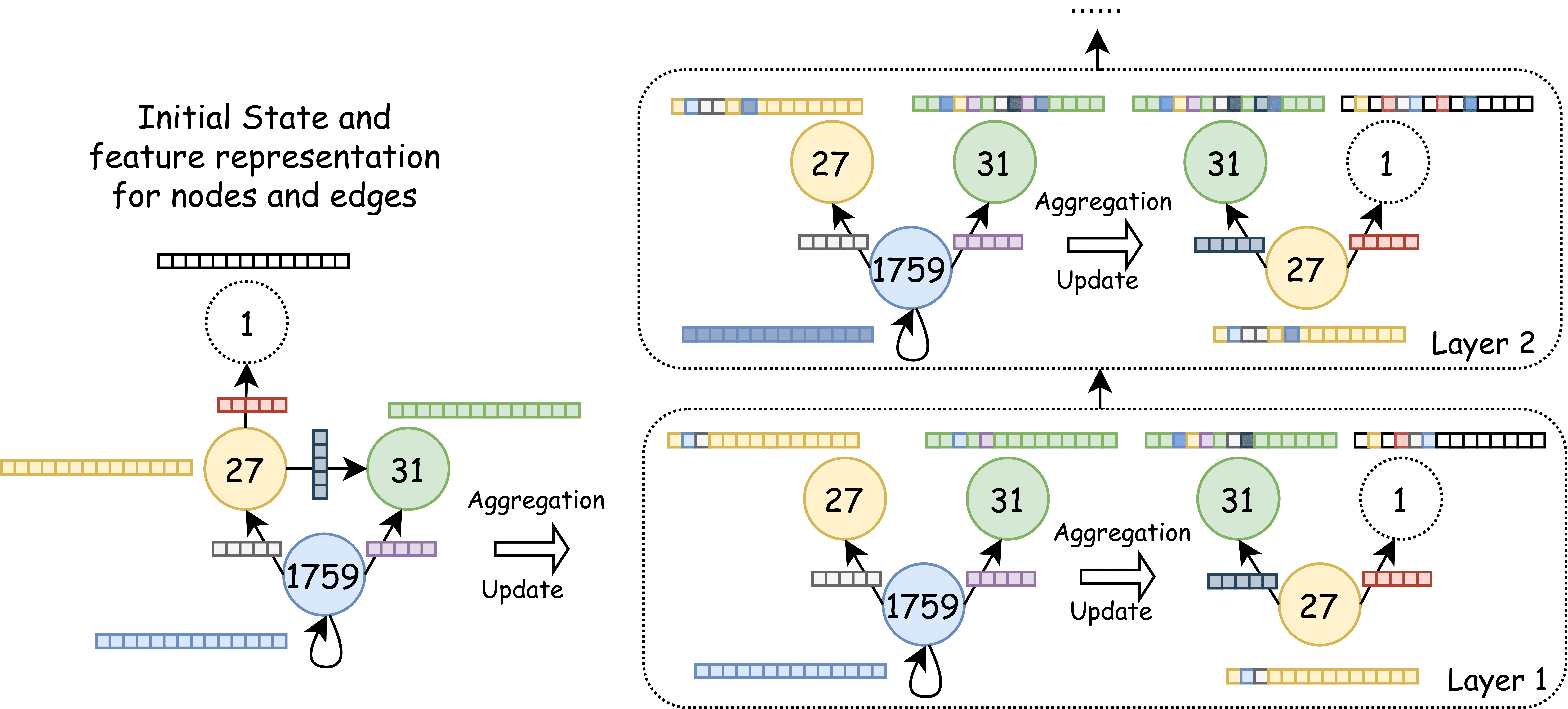}
    \caption{Messages propagate from the target node to its operand nodes}
    \label{fig:mss}
\end{figure}

Figure~\ref{fig:mss} illustrates the message-passing process on a partial SCM decomposition graph with prefix length $k=2$, using the constant $c=1759$ as a running example. In this graph, each node corresponds to an intermediate multiplier appearing in the decomposition (e.g., $1759$, $31$, $27$, $1$), and directed edges encode operand dependencies induced by previously selected SCM operators.

In the figure, the rows of colored squares attached to nodes and edges visualize their associated feature vectors: different colors represent different node or edge features. These features constitute the raw information exchanged during message passing.

At each GNN layer, messages are propagated along directed operand edges. For example, if the constant $1759$ was decomposed, then information flows from node $1759$ toward nodes $31$ and $27$. The edge features encode semantic information, allowing the model to distinguish, for instance, a shifted-left operand from a subtractive one.

Formally, at message-passing layer $k$, the hidden representation of a node $v$ is updated as:

\begin{equation}
h_v^{(k)} =
\gamma^{(k)}\!\left(
h_v^{(k-1)},
\bigoplus_{u \in \mathcal{N}(v)}
\phi^{(k)}\!\left(h_v^{(k-1)}, h_u^{(k-1)}, \mathbf{e}_{vu}\right)
\right),
\end{equation}

where $h_v^{(k)}$ denotes the latent embedding of node $v$, $\mathcal{N}(v)$ is the set of predecessor nodes connected to $v$, and $\mathbf{e}_{vu}$ represents the edge features between nodes $u$ and $v$.

In the SCM setting, this update has a clear semantic interpretation. 
Messages are propagated from a parent node to its child nodes along directed operand edges. 
For example, information flows from the current target constant $1759$ toward its subcomponents $27$ and $31$. 
The node features of $1759$ encode global arithmetic properties such as bit-length, density of set bits, and proximity to powers of two, while the associated edge features specify how this constant is decomposed, including operand roles and shift-related attributes. 

Through message passing, each child node (e.g., $27$ or $31$) receives contextual information describing \emph{why} and \emph{how} it was generated from $1759$. This enables the model to contextualize the local properties of a sub-constant within the broader decomposition objective, such as whether it is expected to be realized via shifts, additions, or subtractions in subsequent steps.

\subsection{Good Rules Generation and Confidence Quantification}

After training the GNN model, we deploy it for inference to predict the optimal operator(s) for decomposing a given constant, as presented in Figure~\ref{workflowI}. Since ML models are inherently probabilistic rather than deterministic, we use the prediction probabilities as confidence scores for each candidate operator. These confidence scores form the basis for extracting actionable pruning rules.

We define the following policy to translate predicted probabilities into symbolic \emph{good rules}:

\begin{enumerate}
    \item \textit{High-Confidence Rule}: If the top-ranked operator’s predicted confidence exceeds that of the second-ranked operator by more than a threshold $\delta$ (set to 0.2), only the top operator is selected as a \emph{good rule} for that decomposition step.
    \begin{equation}
        \text{good\_rule}(c, s, o, \text{conf}) \Leftrightarrow \text{conf}(o) - \max_{o' \neq o} \text{conf}(o') > \delta
    \end{equation}
    where $c$ is the constant, $s$ is the graph state, $o$ is the operator, and $\text{conf}(o)$ is its predicted confidence.
    
    \item \textit{Ambiguity-Aware Multiple Rules}: If the confidence gap between the top two operators is less than or equal to $\delta$, the decision is considered ambiguous and both operators are selected as \emph{good rules} to avoid prematurely pruning potentially optimal branches.
    \begin{equation}
        \begin{aligned}
        \text{good\_rule}(c, s, o_1, \text{conf}_1) \land \text{good\_rule}(c, s, o_2, \text{conf}_2) \Leftrightarrow \\
        |\text{conf}_1 - \text{conf}_2| \leq \delta \land o_1, o_2 \in \text{top\_two\_predictions}
        \end{aligned}
    \end{equation}
\end{enumerate}

This policy produces a set of recommended operators for a given constant. Conversely, operators whose confidence consistently falls below a minimum threshold \(\epsilon\) are compiled as \emph{no-good rules} and excluded from search to prune unpromising branches.

\subsection{Integrating Rules into the SAT Encoding}

 For each constant $c$, the GNN model outputs a probability distribution over the SCM operator types. These probabilities are translated into symbolic predicates, producing an interface that contains facts of the form $\texttt{op}(c, o)$, indicating that operator $o$ is recommended for decomposing constant $c$. The identified \textit{good} rules are then integrated into the \textit{DP-mink} encoding by constraining the predicate \texttt{op/2}. Concretely, instead of non-deterministically exploring all three operators at every decomposition step, the solver consults the identified \textit{good} rules and restricts the operator domain accordingly.

The resulting rule-guided DP encoding is shown below.

\noindent
\rule{\textwidth}{1pt}
{\footnotesize
\begin{verbatim}
import ordset.
import Good_Rules.

table (+, mmin, -)
dpscm(1, Size, Plan) =>
    Size = 0, Plan = [].
    
dpscm(C, Size, Plan) =>
    op(C, OP),                          % Good Rules as an additional constraint   
    scm(C, Size, Plan, OP)
    
scm(C, Size, Plan, OP) =>
    if OP = SPLUS then
        split_pp(C, C1, C2, S),           % SPLUS: C = (C1 << S) + C2
        OP = $splus(C, C1, C2, S)
    elif OP = SMINUSS then
        split_pn(C, C1, C2, S),           % SMINUS: C = (C1 << S) - C2
        OP = $sminus(C, C1, C2, S)
    else
        split_np(C, C1, C2, S),           % MINUSS: C = C1 - (C2 << S)
        OP = $minuss(C, C1, C2, S),  
        ......
\end{verbatim}
}
\rule{\textwidth}{1pt}


\section{Experiments and Results Evaluation}
\label{exp}

In this section, we evaluate the effectiveness of the proposed ML-DP framework. 
Our evaluation focuses on four aspects of the SCM encoding process: encoding quality, encoding time, memory consumption, and search scalability. 

We compare four methods:

\begin{itemize}
    \item Baseline: a direct shift-and-add encoding without decomposition optimization;
    \item Min-k: the \textit{min-k} encoding strategy that minimizes the number of additions as used in~\cite{bierlee2024single} using MiniZinc;
    \item DP: the DP method implemented in Picat~\cite{zhoupicat2026};
    \item ML-DP: the proposed ML-guided DP approach with learned good rules.
\end{itemize}

\subsection{Dataset}
\label{dataset}

The training set is constructed from two sources of constants. 
First, we include all odd constants in the range $1$--$65{,}535$, which were used in previous SCM studies~\cite{bierlee2024single,zhoupicat2026}. Second, to expose the model to larger numerical structures, we randomly sample 6,000 odd constants for each bit-length from 17 to 32. In total, the training set contains 128,767 constants, with training labels generated by the Picat DP solver. For each constant and its recipe, a sequence of graphs as shown in Figure~\ref{gscm} are generated as training instances.

To evaluate generalization, the test set consists of previously unseen constants with bit-lengths from 17 to 32. For each bit-length in this range, we randomly generate 1,000 odd constants, resulting in a total of 16,000 test instances. 

Although the proposed method can naturally be applied to larger constants, in this experiment, we restrict our evaluation to the 17--32 bit range. The framework can be readily applied to larger bit-widths (e.g., 64 or 128 bits) by generating additional training data, without modifying the underlying symbolic solver.

\subsection{Encoding Time and Encoding Quality}

Encoding time is the dimension in which our method demonstrates the most substantial advantage. 
Encoding quality is measured by the number of additions (\#adds), which directly reflects the size and complexity of the resulting arithmetic circuit. 

The Min-k method times out in our experiments, failing to solve all test instances within the 24-hour time limit. Therefore, we report results only for the remaining three methods.

As shown in Table~\ref{tab:encodingk}, the Baseline method achieves the fastest runtime but produces encodings with the largest number of additions. In contrast, the runtime of DP grows rapidly with increasing bit-length, reaching over 4.6 seconds at 32 bits. Our ML-DP solver consistently operates within milliseconds, even for the largest constants tested. At 32 bits, ML-DP achieves a speedup of approximately $87\times$ over DP. More generally, across the range of 17--32 bits, we observe speedups ranging from about $10\times$ to over $100\times$. This improvement stems from the learned good rules, which significantly reduce backtracking and operator branching during DP recursion. Importantly, this acceleration is achieved without sacrificing symbolic correctness of the encodings produced, while maintaining near-optimal quality in practice.

In terms of encoding quality, Table~\ref{tab:encodingk} shows that the DP method consistently produces the smallest number of additions across all bit-lengths. Our ML-DP method yields slightly larger encodings but still substantially improves over the Baseline. For example, for 32 bits, the Baseline requires on average 18 additions, while DP achieves 7.26 additions and ML-DP produces 8.85 additions. Thus, compared to the Baseline, ML-DP reduces the number of additions by approximately 9 operations. At the same time, it introduces only about 1.6 additional operations relative to DP, corresponding to roughly one to two extra additions in the resulting recipe.

Overall, these results demonstrate a favorable trade-off: ML-DP achieves near-optimal encoding quality while providing orders-of-magnitude improvements in encoding time.

\begin{table}[htbp]
\centering
\label{tab:combinedtables}

\begin{minipage}[t]{0.45\textwidth}
\centering
\caption{Comparison of average encoding time (in seconds) and encoding quality (\# additions)}
\label{tab:encodingk}
\scriptsize
\vspace{0.5em}
\begin{tabular}{@{}lrcrcrc@{}}
\toprule
\toprule
\multicolumn{1}{c}{\multirow{2}{*}{Length}} & 
\multicolumn{2}{c}{Baseline} & 
\multicolumn{2}{c}{DP} & 
\multicolumn{2}{c}{ML-DP} \\
\cmidrule(lr){2-3} \cmidrule(lr){4-5} \cmidrule(lr){6-7}
 & time & \#adds & time & \#adds & time & \#adds \\
\midrule
17      & 0.001 &  10   & 0.025  & 4.452 & 0.000 & 5.226 \\
18     & 0.001 &   9  & 0.037 & 4.663 & 0.000 & 5.525 \\
19     & 0.001 &  8   & 0.056 & 4.857 & 0.001 & 5.779 \\
20     & 0.001 &  12   & 0.085 & 5.059 & 0.001 & 6.059 \\
21    & 0.001 &  14   & 0.123 & 5.267 & 0.001 & 6.317 \\
22    & 0.001 &  15   & 0.177 & 5.451 & 0.002 & 6.518 \\
23    & 0.001 &  12   & 0.258 & 5.613 & 0.002 & 6.723 \\
24   & 0.001 &  15   & 0.376 & 5.786 & 0.003 & 7.006 \\
25   & 0.001 &  12   & 0.521 & 6.007 & 0.004 & 7.257 \\
26   & 0.001 &  16   & 0.717 & 6.180 & 0.006 & 7.500 \\ 
27   & 0.001 &  12   & 0.982 & 6.375 & 0.008 & 7.714 \\
28  & 0.001 &  19   & 1.388 & 6.581 & 0.011 & 7.973 \\
29  & 0.001 &  15   & 1.895  & 6.738 & 0.016 & 8.228 \\
30  & 0.001 &  17   & 2.673 & 6.918 & 0.023 & 8.399 \\
31 & 0.001 &  18   & 3.600 & 7.099 & 0.033 & 8.610 \\
32 & 0.001 &  18   & 4.634 & 7.260 & 0.053 & 8.846 \\
\bottomrule
\bottomrule
\end{tabular}
\end{minipage}
\hfill 
\begin{minipage}[t]{0.45\textwidth}
\centering
\caption{Comparison of average memory consumption (in MB) and branching times}
\label{tab:encodingm}
\scriptsize
\vspace{0.5em}
\begin{tabular}{@{}lrcrcrc@{}}
\toprule
\toprule
\multicolumn{1}{c}{\multirow{2}{*}{Length}} & 
\multicolumn{2}{c}{DP} & 
\multicolumn{2}{c}{ML-DP} \\
\cmidrule(lr){2-3} \cmidrule(lr){4-5} 
 & memory & branch & memory & branch  \\
\midrule
17      & 1.634 &  4863   & 0.066  & 135  \\
18     & 2.295 &   6317  & 0.087 & 155  \\
19     & 3.089 &  7988   & 0.118 & 185 \\
20     & 4.335 &  10104   & 0.146 & 215  \\
21    & 5.742 &  12360   & 0.198 & 252  \\
22    & 7.583 &  15055   & 0.258 & 288 \\
23    & 9.866 &  18421   & 0.322 & 318  \\
24   & 12.847 &  22144   & 0.399 & 357  \\
25   & 17.070 &  26386   & 1.536 &  409  \\
26   & 21.581 &  31228   & 0.687 &  465  \\ 
27   & 27.926 &  36710   & 0.858 &  519  \\
28  & 35.688 &  43059   & 1.070 &  561  \\
29  & 45.772 &  50258   & 1.411  &  627  \\
30  & 55.727 &  57884   & 1.676 &  682 \\
31 & 69.843 &  66487   & 2.131 &  745  \\
32 & 87.553 &  76483   & 2.688 &  825  \\
\bottomrule
\bottomrule
\end{tabular}
\end{minipage}
\end{table}

\subsection{Memory Consumption and Branching Behavior}

We evaluate memory consumption and branching behavior for methods that rely on dynamic programming with tabling. The Baseline and Min-k methods do not employ DP-based search or maintain intermediate DP tables; therefore, memory and branching metrics are not applicable to these methods.

Memory usage reflects the size of the DP tables maintained during search. 
As shown in Table~\ref{tab:encodingm}, DP exhibits steep memory growth as the bit-length increases, exceeding 87 MB on average at 32 bits. In contrast, our ML-DP approach maintains consistently low memory usage, consuming only 2.7 MB on average at 32 bits. This corresponds to a reduction of approximately 97\% in memory consumption. This reduction is a direct consequence of the smaller DP tables induced by rule-guided pruning, which prevents the solver from exploring unproductive branches. Such efficiency is particularly important for deployment in resource-constrained environments, including edge-device pipelines and hardware-oriented synthesis settings.

To further understand the scalability of the proposed method, we examine the number of branchings explored during search. As shown in Table~\ref{tab:encodingm}, DP method exhibits rapid growth in branching, exceeding 76{,}000 branches at 32 bits. In contrast, our ML-DP approach maintains fewer than 1{,}000 branches even at 32 bits, representing an order-of-magnitude reduction in search complexity.

These results demonstrate that the learned good rules effectively reshape the symbolic search tree, transforming an otherwise combinatorial explosion into a more controlled and predictable exploration process. Importantly, although the guidance is learned from data, the resulting pruning is not purely heuristic in the conventional sense. Instead, it is explicitly compiled into symbolic constraints, making the pruning process explainable and preserving the soundness of the underlying symbolic search.

\subsection{GNN Hyperparameter Analysis}
\label{ablation}

To assess the sensitivity of the proposed GNN model to architectural choices, we conduct a hyperparameter ablation study over three key factors: hidden 
dimension $d \in \{128, 256, 512\}$, number of message-passing 
layers $N \in \{3, 4, 5\}$, and number of attention heads 
$H \in \{4, 8\}$. All variants are trained under identical 
conditions (batch size 256, initial learning rate $2\times10^{-4}$, 
cosine warm-restart schedule, up to 800 epochs with early stopping 
based on validation loss with patience 20) and evaluated on the same 
validation set. Among all configurations, the model with $d=256$, $N=4$, and $H=4$ achieves the best performance (98.52\%). This configuration is therefore adopted in all subsequent experiments.

\begin{table}[ht]
\centering
\caption{Hyperparameter ablation study of the GNN.}
\scriptsize 
\label{tab:ablation}
\vspace{0.5em}
\begin{minipage}[t]{0.48\textwidth}
\centering
\begin{tabular}{cccc}
\toprule
Hidden Dim & Layers & \#Heads & Val Acc (\%) \\
\midrule
128 & 3 & 4 & 98.47 \\
256 & 3 & 4 & 98.42 \\
512 & 3 & 4 & 98.48 \\
\midrule
128 & 4 & 4 & 98.46 \\
\textbf{256} & \textbf{4} & \textbf{4} & \textbf{98.52} \\
512 & 4 & 4 & 98.46 \\
\midrule
128 & 5 & 4 & 98.43\\
256 & 5 & 4 & 98.50 \\
512 & 5 & 4 & 98.50 \\
\bottomrule
\end{tabular}
\end{minipage}
\hfill
\begin{minipage}[t]{0.48\textwidth}
\centering
\begin{tabular}{cccc}
\toprule
Hidden Dim & Layers & \#Heads & Val Acc (\%)  \\
\midrule
128 & 3 & 8 & 98.50\\
256 & 3 & 8 & 98.47 \\
512 & 3 & 8 & 98.45\\
\midrule
128 & 4 & 8 &  98.45 \\
256 & 4 & 8 & 98.51 \\
512 & 4 & 8 &  98.48\\
\midrule
128 & 5 & 8 & 98.43\\
256 & 5 & 8 &  98.51\\
512 & 5 & 8 & 98.48\\
\bottomrule
\end{tabular}
\end{minipage}
\end{table}

As shown in Table~\ref{tab:ablation}, the model achieves consistently high accuracy (above 98\%) across a wide range of configurations, demonstrating strong robustness to hyperparameter variations. This indicates that the effectiveness of the method is primarily driven by the underlying design rather than careful tuning.

\section{Related Work}
\label{relatedwork}

The SCM problem has been extensively studied in hardware design and shown to be NP-hard for minimizing additions and subtractions~\cite{de2023multiplication,cappello2003some}. Its generalization, Multiple Constant Multiplication (MCM), has been formulated as a combinatorial optimization problem, notably by Gustafsson as a Steiner hypertree on directed hypergraphs, solved optimally via ILP with practical constraints such as adder depth and fan-out~\cite{gustafsson2008towards}. To further reduce adder cost in multiplierless realizations, recursive common subexpression elimination methods were proposed~\cite{macleod2004common}.
Subsequent ILP-based formulations improved hardware awareness by supporting multi-input adders, low-power optimization, and error-aware cost models, but still suffer from scalability limitations~\cite{gustafsson2008towards,kumm2018optimal,garcia2023toward}. As a result, SAT-based approaches emerged as a powerful alternative, enabling optimal or provably minimal SCM/MCM synthesis even for large bit-width constants and general linear transforms~\cite{lagoon2020deriving}. Bierlee \textit{et al.} further advanced SAT-based SCM synthesis by introducing SAT-specific encodings that minimize full/half adders and by providing a reusable library of optimal SCM recipes to accelerate solving~\cite{bierlee2024single}. In parallel, approximate methods have been proposed to trade optimality for scalability in practical digital signal processing and filter design~\cite{aksoy2014exact,lefevre2001multiplication}.

\section{Conclusion}
\label{conclusion}

In this paper, we presented a neuro-symbolic framework for accelerating SAT encodings of the SCM problem. By integrating a GNN with a DP solver, the approach learns \emph{good rules} that guide operator selection during decomposition, reducing the search space while maintaining near-optimal encoding quality. Experimental results show substantial improvements in encoding time, memory usage, and branching, demonstrating that learning-guided pruning can significantly improve the scalability of symbolic search.

More broadly, this work highlights the potential of combining ML with declarative and logic-based frameworks. Rather than replacing symbolic reasoning, the learned model complements it by improving search efficiency. This paradigm may extend to other combinatorial optimization and SAT-encoding problems.

Future work includes extending the approach to the \textit{min-a} objective, exploring richer rule representations, and applying the framework to other symbolic decomposition tasks.

\nocite{*}
\bibliographystyle{eptcs}
\bibliography{generic}
\end{document}